\documentclass[sigconf,preprint]{acmart}
\AtBeginDocument{%
  }

\setcopyright{acmlicensed}
\copyrightyear{2025}
\acmYear{2025}
\acmDOI{XXXXXXX.XXXXXXX}

\acmISBN{978-1-4503-XXXX-X/2018/06}

\usepackage{hyperref}

\usepackage{amssymb}

\begin{document}

\title{BasketHAR: A Multimodal Dataset for Human Activity Recognition and Sport Analysis in Basketball Training Scenarios}



\author{Xian Gao}
\affiliation{%
  \institution{Shanghai Jiao Tong University}
  \city{Shanghai}
  \country{China}}
\email{gaoxian@sjtu.edu.cn}

\author{Haoyue Zhang}
\affiliation{%
  \institution{Shanghai Jiao Tong University}
  \city{Shanghai}
  \country{China}}
  
\author{Zongyun Zhang}
\affiliation{%
  \institution{Shanghai Jiao Tong University}
  \city{Shanghai}
  \country{China}}

\author{Jiacheng Ruan}
\affiliation{%
  \institution{Shanghai Jiao Tong University}
  \city{Shanghai}
  \country{China}}

\author{Ting Liu}
\affiliation{%
  \institution{Shanghai Jiao Tong University}
  \city{Shanghai}
  \country{China}}

\author{Yuzhuo Fu}
\authornote{Corresponding author.}
\affiliation{%
  \institution{Shanghai Jiao Tong University}
  \city{Shanghai}
  \country{China}}


\renewcommand{\shortauthors}{Gao et al.}


%



\renewcommand\footnotetextcopyrightpermission[1]{}
\settopmatter{printacmref=false} 


\begin{abstract}

Human Activity Recognition (HAR) involves the automatic identification of user activities and has gained significant research interest due to its broad applicability. Most HAR systems rely on supervised learning, which necessitates large, diverse, and well-annotated datasets. However, existing datasets predominantly focus on basic activities such as walking, standing, and stair navigation, limiting their utility in specialized contexts like sports performance analysis. To address this gap, we present BasketHAR, a novel multimodal HAR dataset tailored for basketball training, encompassing a diverse set of professional-level actions. BasketHAR includes comprehensive motion data from inertial measurement units (accelerometers and gyroscopes), angular velocity, magnetic field, heart rate, skin temperature, and synchronized video recordings. We also provide a baseline multimodal alignment method to benchmark performance. Experimental results underscore the dataset’s complexity and suitability for advanced HAR tasks. Furthermore, we highlight its potential applications in the analysis of basketball training sessions and in the generation of specialized performance reports, representing a valuable resource for future research in HAR and sports analytics. The dataset are publicly accessible at \href{https://huggingface.co/datasets/Xian-Gao/BasketHAR}{\color{blue}{https://huggingface.co/datasets/Xian-Gao/BasketHAR}} licensed under Apache License 2.0.

\end{abstract}

\maketitle

\section{Introduction}
\label{sec:intro}

Advancements in wearable devices and sensor technologies have significantly expanded the applications of Human Activity Recognition (HAR) in areas such as health monitoring, sports training, and human-computer interaction. HAR involves identifying specific human actions by analyzing motion signals from targeted body regions (e.g., acceleration, angular velocity, joint angles), typically framed as a classification task. Sensor-based HAR, in particular, utilizes multimodal signals—such as accelerometer and gyroscope data—alongside advanced algorithms, including deep learning, to enable accurate and efficient activity recognition.

In recent years, numerous open-source HAR datasets have been developed, providing essential resources for academic research. The UCI HAR dataset \cite{anguitaPublicDomainDataset2013} focuses on daily activities like walking and stair navigation, while the WISDM dataset \cite{kwapiszActivityRecognitionUsing2011b} captures common behaviors using smartphone accelerometers. The Opportunity dataset \cite{chavarriagaOpportunityChallengeBenchmark2013} employs multiple sensors to collect motion data in complex settings, targeting industrial and health-related applications. In the sports domain, datasets such as KU-HAR \cite{sikderKUHAROpenDataset2021} feature relatively static activities like yoga and gymnastics. However, existing datasets largely emphasize static or simple dynamic actions and lack detailed annotations of complex movements typical in athletic training, such as dribbling, passing, and shooting. Furthermore, most current datasets rely solely on accelerometer and gyroscope data, whereas in-depth analysis of sport training requires multimodal signals, including angular velocity, magnetic field, and heart rate. Additionally, the predominance of laboratory-based data collection limits the ability to capture the nuanced and dynamic characteristics of real-world sports environments.

To address the identified research gap, this paper presents a multimodal HAR dataset tailored for basketball training scenarios. The dataset captures a wide range of signals, including acceleration, gyroscopic data, angular velocity, magnetic field, heart rate, skin surface temperature, and synchronized video recordings. This multimodal approach enables detailed representation of both the dynamic movements and the physiological states associated with basketball training. The dataset includes annotated instances of key basketball actions—such as dribbling, passing, and shooting—collected in real-world training environments to ensure ecological validity and broad applicability.

The main contributions of this paper are as follows:  
\begin{enumerate}
    \item We introduce the first multimodal HAR dataset tailored to basketball training scenarios, offering a novel data resource for action recognition research in the sports domain.
    \item For the first time, heart rate and skin surface temperature signals are incorporated to provide a more comprehensive perspective on physical activity. This enables new avenues for investigating the relationship between physiological states and action classification, as well as for exploring multimodal fusion strategies.
    \item We propose a baseline method based on multimodal alignment and experimentally demonstrate the effectiveness of multimodal signals in recognizing complex and dynamic actions. This establishes a baseline for activity recognition and automated analysis report generation in basketball training contexts.
\end{enumerate}
\section{Related Works}

\subsection{Human Activity Recognition}

Human Activity Recognition (HAR) entails identifying and interpreting human behaviors, often using motion sensors such as Inertial Measurement Units (IMUs) that record accelerometer and gyroscope data. Owing to the ease of acquiring motion signals, IMUs are widely adopted in HAR applications on smartphones and wearable devices. Deep learning models, including CNNs \cite{zengConvolutionalNeuralNetworks2014,ronaoHumanActivityRecognition2016,huangShallowConvolutionalNeural2021}, RNNs \cite{muradDeepRecurrentNeural2017,inoueDeepRecurrentNeural2018}, and LSTMs \cite{guanEnsemblesDeepLSTM2017,ashryCHARMDeepContinuousHuman2020}, are commonly employed for feature extraction and classification of these signals. Recent approaches enhance IMU representations through large-scale pre-training and contrastive learning by aligning them with other modalities, resulting in generalized, transferable features applicable to diverse downstream HAR tasks \cite{moonIMU2CLIPMultimodalContrastive2022,girdharImageBindOneEmbedding2023a}. These pre-trained models exhibit robust generalization, performing well on complex, real-world motion data. Nevertheless, HAR performance is highly dependent on dataset diversity. This study introduces a HAR dataset specific to basketball training, incorporating IMU signals, angular velocity, magnetic field data, and additional sensor inputs, thus offering a more comprehensive and varied dataset for training and evaluation.

\subsection{Real-world HAR Dataset}

The widespread integration of sensors such as accelerometers and gyroscopes in consumer devices like smartphones has led to the development of numerous publicly available HAR datasets. Notable examples include UCI HAR \cite{anguitaPublicDomainDataset2013}, WISDM \cite{kwapiszActivityRecognitionUsing2011b}, USC-HAD \cite{zhangUSCHADDailyActivity2012a}, and KU-HAR \cite{sikderKUHAROpenDataset2021}, which primarily capture accelerometer and gyroscope data from daily activities. Garcia-Gonzalez et al. \cite{garcia-gonzalezPublicDomainDataset2020} extended this by incorporating magnetic field and GPS data. However, smartphone-based data collection presents challenges, including limited sensor precision, lower sampling rates, and restricted sensor variety. Additionally, attaching smartphones to the body can disrupt natural movement, leading to discrepancies between recorded and actual motion. To address these limitations, datasets like UMAFall \cite{casilariUMAFallMultisensorDataset2017}, Opportunity Challenge \cite{chavarriagaOpportunityChallengeBenchmark2013}, and DU-MD \cite{sahaDUMDOpenSourceHuman2018} utilize wearable sensors, offering higher accuracy and broader signal types. Nonetheless, these datasets focus mainly on routine activities (e.g., walking, sitting, standing) and lack domain-specific actions. Since dataset diversity directly impacts the range of activities a model can recognize, this study introduces a dataset encompassing both general and basketball-specific movements, designed to support HAR in both everyday and sports training contexts.

\section{Dataset Construction}

In this section, we present the methodology for constructing the dataset, which includes data collection and annotation processes. Upon completion of the dataset construction, we provide detailed information regarding its contents.

\subsection{Data Collection}
\subsubsection{Settings}

\begin{figure}[!t]
\centerline{\includegraphics[width=0.85\columnwidth]{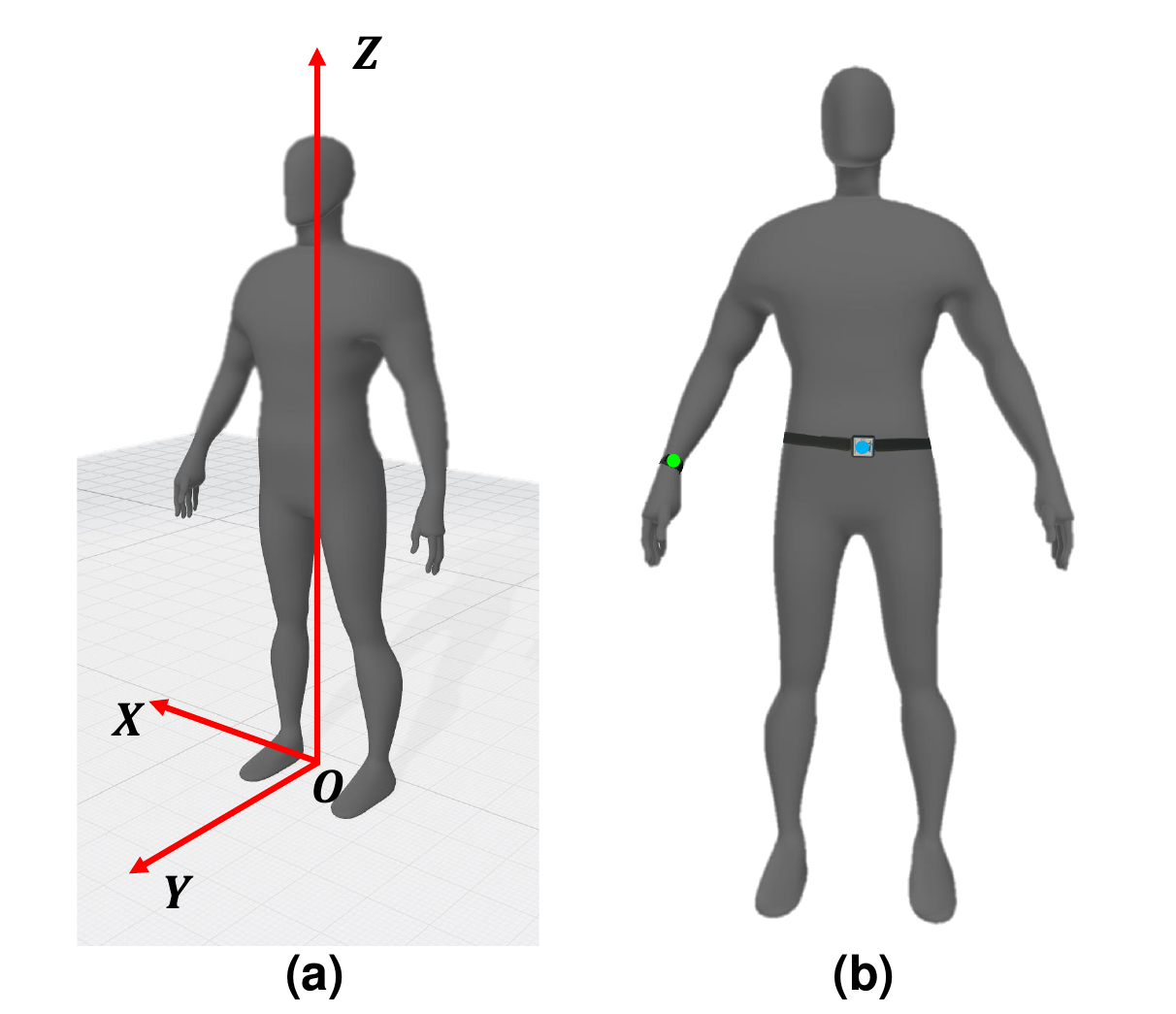}}
\caption{\textbf{(a) Axes of Motion Relative to User.}  \textbf{(b) Sensor Placement} Sensors for collecting acceleration, gyroscope data, angle, and magnetic field were mounted on the front center of the volunteer's waist (the \textcolor[RGB]{0,176,240}{blue} point), while sensors for monitoring heart rate and body surface temperature were worn on the wrist of the volunteer's dominant hand (the \textcolor{green}{green} point).}
\label{axis}
\Description{\textbf{(a) Axes of Motion Relative to User.}  \textbf{(b) Sensor Placement} Sensors for collecting acceleration, gyroscope data, angle, and magnetic field were mounted on the front center of the volunteer's waist (the \textcolor[RGB]{0,176,240}{blue} point), while sensors for monitoring heart rate and body surface temperature were worn on the wrist of the volunteer's dominant hand (the \textcolor{green}{green} point).}
\end{figure}

\begin{figure*}[!t]
\centerline{
\includegraphics[width=0.9\textwidth]{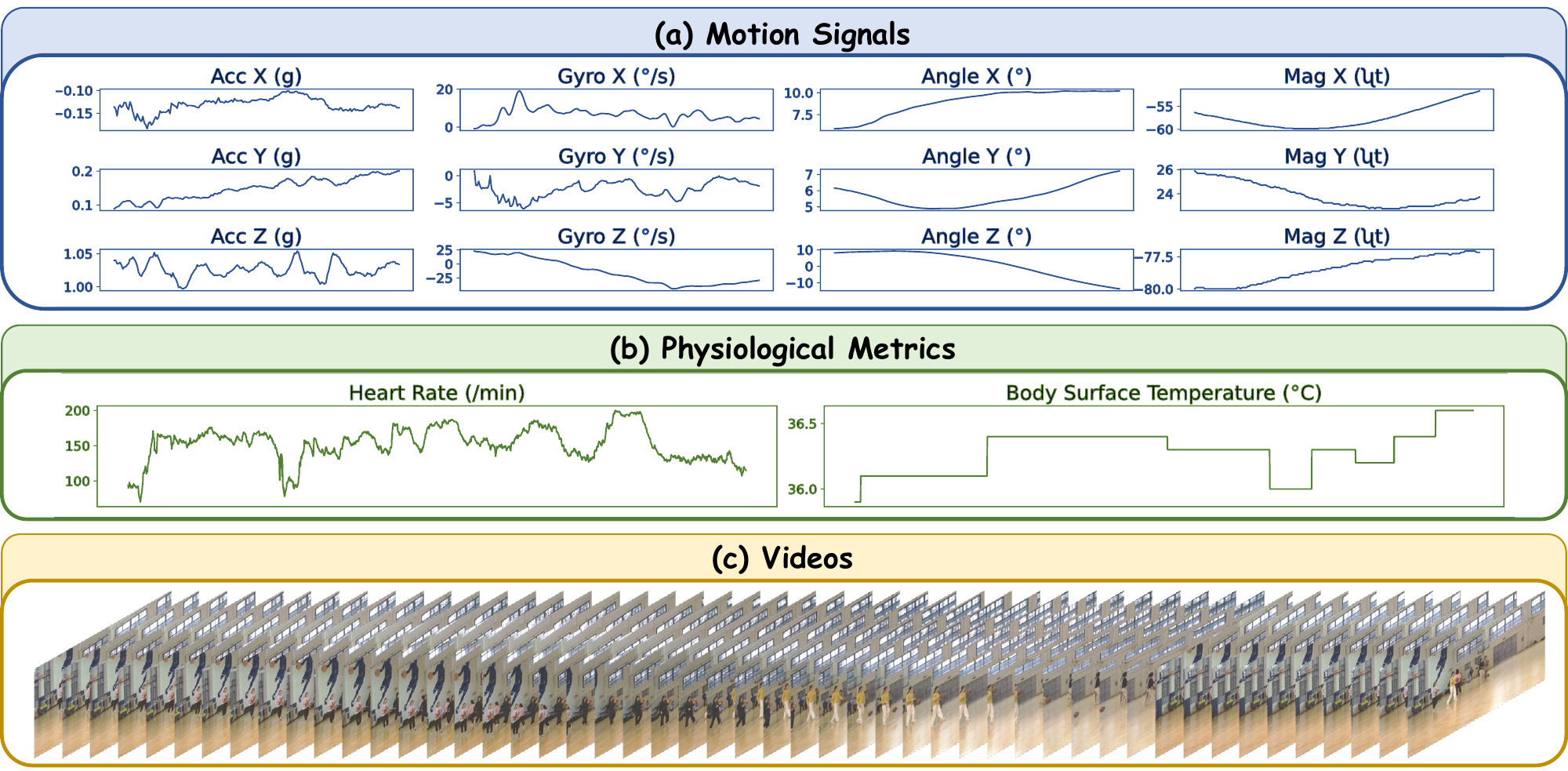}}
\caption{\textbf{Dataset Overview.} The BasketHAR dataset encompasses three modalities of data collected synchronously during physical activities: motion signals, physiological metrics, and activity videos.}
\label{signal}
\Description{\textbf{Dataset Overview.} The BasketHAR dataset encompasses three modalities of data collected synchronously during physical activities: motion signals, physiological metrics, and activity videos.}
\end{figure*}

\begin{table*}[!t]
\centering
\caption{Description of the activity classes in BasketHAR dataset.}
\small
\begin{tabular}{cccc}
\hline
Class Name                      & Class ID & Performed Activity                                                   & \begin{tabular}[c]{@{}c@{}}\# Collected\\ Samples\end{tabular} \\ \hline
Sit                             & 0        & To rest on a chair or floor with the body supported by the buttocks. & 497                                                                      \\
Stand                           & 1        & To maintain an upright position on the feet.                         & 2165                                                                     \\
Warm-up Exercises               & 2        & To perform light exercises to prepare the body for intense activity. & 582                                                                     \\
Walk                            & 3        & To move at a regular pace by lifting and setting down each foot.     & 931                                                                     \\
Run                             & 4        & To move quickly on foot by taking longer strides than walking.       & 65                                                                      \\
Dribble Run                     & 5        & To run while bouncing the basketball continuously.                   & 230                                                                      \\
Hold Ball Standing              & 6        & To stand while holding a basketball without moving.                  & 1882                                                                     \\
Bounce the Ball                 & 7        & To repeatedly push the ball downward so it rebounds off the floor.   & 3100                                                                     \\
Travel (with Ball)              & 8        & To walk while holding the basketball, violating basketball rules.    & 626                                                                      \\
Shoot the Ball                  & 9       & To throw the basketball toward the hoop in an attempt to score.      & 621                                                                      \\
Pass on the Run                 & 10       & To pass the basketball to a teammate while running.                  & 94                                                                      \\
Low Dribble (Alternating Hands) & 11       & To dribble the ball close to the ground, alternating hands.          & 83                                                                      \\
Low Dribble (Right Hand)        & 12       & To dribble the ball close to the ground using only the right hand.   & 235                                                                      \\
Low Dribble (Left Hand)         & 13       & To dribble the ball close to the ground using only the left hand.    & 124                                                                      \\ \hline
\end{tabular}
\label{dataset}

\end{table*}

During the data collection process, sensors measuring acceleration, gyroscope data, angle, and magnetic field were affixed to the front center of the volunteer's waist. Prior to the experiment, the sensors were calibrated in a stationary state to establish reference zero points and coordinate axis orientations (Figure \ref{axis}(a)). Sensors for heart rate and body surface temperature monitoring were worn on the wrist of the volunteer's dominant hand (Figure \ref{axis}(b)). Subsequently, the volunteers participated in a 90-minute basketball training session on a standard basketball court, during which the sensors remained continuously active, sampling and transmitting data at a predefined frequency. A host computer, used for data collection, was positioned at the side of the basketball court and maintained within the same local area network as the sensors throughout the session. To facilitate subsequent data annotation, video recordings of the basketball training session were captured simultaneously with the collection of sensor signals.

\subsubsection{Collection Methods}

We used the MPU9250, a nine-axis sensor manufactured by InvenSense Inc.
, to collect the required data, including accelerometer, gyroscope, angle, and magnetic field data. This chip integrates an accelerometer, gyroscope, and magnetometer, providing three-dimensional acceleration, angular velocity, and magnetic field intensity data. The MPU9250 is embedded in the WT901 SoC, and the sensor data is transmitted to the host computer via a UART interface. During data collection, the sensor sampling rate was set to 200~Hz. Heart rate and surface temperature data were collected using the nRF51822 SoC, which integrates the Silicon Labs Si1141 sensor
. These data were transmitted to the host computer via Bluetooth, with a sampling rate of 1~Hz.

\begin{table*}[!t]
\centering
\caption{Comparison with other HAR datasets.}
\small
\begin{tabular}{c|ccccc}
\hline
            Datasets                 & Sensors                               & \begin{tabular}[c]{@{}c@{}}Activity \\ Type\end{tabular} & \# Activities & \#  Samples & Sampling Rate (Hz) \\ \hline
UCI HAR       \cite{anguitaPublicDomainDataset2013}                    & Acc. and gyro.                        & Daily                                                    & 6                                                               & 10,299                                                       & 50                 \\
WISDM       \cite{kwapiszActivityRecognitionUsing2011b}                      & Acc. and gyro.                        & Daily                                                    & 6                                                               & 5,424                                                        & 20                 \\
USC-HAD    \cite{zhangUSCHADDailyActivity2012a}                      & Acc. and gyro.                        & Daily                                                    & 12                                                              & 840                                                          & 100                \\
WARD   \cite{yangDistributedRecognitionHuman2009}                         & Acc. and gyro.                        & Daily                                                    & 13                                                              & 1,298                                                        & 30                 \\
UMAFall      \cite{casilariUMAFallMultisensorDataset2017}                        & Acc., gyro., and mag.                 & Daily                                                    & 11                                                              & 531                                                          & 200                \\
HASC Challenge    \cite{kawaguchiHASCChallengeGathering2011}             & Acc.                                  & Daily                                                    & 6                                                               & 6,700                                                        & 100                \\
DU-MD        \cite{sahaDUMDOpenSourceHuman2018}                  & Acc. and gyro.                        & Daily                                                    & 10                                                              & 5,000                                                        & 30                 \\
UCI HAPT  \cite{reyes-ortizHumanActivityRecognition2014}                      & Acc. and gyro.                        & Daily                                                    & 12                                                              & 10,929                                                       & 50                 \\
Garcia-Gonzalez et al.  \cite{garcia-gonzalezPublicDomainDataset2020}      & Acc., gyro., mag., and GPS   & Daily                                                    & 4                                                               & 29,126,810                                                   & Varies             \\
KU-HAR   \cite{sikderKUHAROpenDataset2021}                     & Acc. and gyro.                        & Daily                                                    & 18                                                              & 20,750                                                       & 100                \\ \hline
\textbf{BasketHAR (ours)} & \textbf{\begin{tabular}[c]{@{}c@{}}Acc., gyro., angle, and mag.\\ Heart rate and temperature\\ Video\end{tabular}} & \textbf{\begin{tabular}[c]{@{}c@{}}Daily and \\ domain-specific\end{tabular}}                      & 14                                                     & 14,044                                                       & \textbf{200}       \\ \hline
\end{tabular}

\label{dataset comparison}
\end{table*}

\subsection{Data Annotation}

We annotate the collected motion signals by referencing the synchronized video recordings, which serve as the ground truth for labeling. The candidate activity labels are derived from technical movements specified by professional basketball coaches during training sessions. Table~\ref{dataset} lists all candidate labels along with their corresponding action descriptions. 

All annotation work is performed by university students with proficient basketball skills. Each data label is verified by at least two annotators to ensure accuracy, and the start and end points of each activity segment are determined by averaging the annotations from at least two individuals. To protect personal privacy, all facial appearances in the dataset have been blurred and manually reviewed.

\subsection{Dataset Details}

Figure \ref{signal} illustrates examples of the data contained in the dataset. The BasketHAR dataset includes three modalities of data collected synchronously during physical activities: motion signals, physiological metrics, and activity videos. The motion signals comprise accelerometer, gyroscope, angle, and magnetic field data. Among these, accelerometer, gyroscope, and angle signals can be utilized for human activity recognition tasks to analyze the types of actions performed by the wearer. The magnetic field signal, on the other hand, represents the wearer’s positional information and can be used to analyze their activity range. The physiological metrics include the wearer’s heart rate and the surface temperature of the body at the sensor placement sites. These metrics enable the analysis of the wearer’s exercise intensity and physical workload. The video modality provides full-length tracking footage of the wearer’s activities, which was primarily used for data annotation but can also serve further purposes such as behavior understanding and pose estimation tasks. Table \ref{dataset comparison} presents a comparison of our proposed dataset with other publicly available HAR datasets. The BasketHAR dataset encompasses a greater number of activity categories and offers advantages in terms of the richness of data modalities and the sampling rate.

\section{Method}

\begin{figure}
    \centering
    \includegraphics[width=\linewidth]{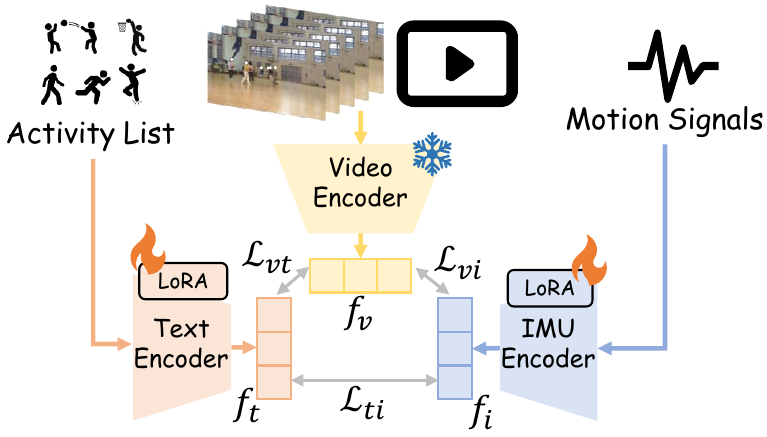}
    \caption{The proposed multimodal alignment approach as a baseline method.}
    \Description{The proposed multimodal alignment approach as a baseline method.}
    \label{fig:method}
\end{figure}

In this section, we propose a baseline method based on multimodal alignment approach (Figure~\ref{fig:method}), designed to fully exploit the multimodal data provided in our dataset and enhance the model's performance on the HAR task.

We perform LoRA fine-tuning \cite{huLoRALowRankAdaptation2021a} on the ImageBind model \cite{girdharImageBindOneEmbedding2023a}, a large-scale pre-trained multimodal alignment model. Since ImageBind aligns various modalities through the visual modality during its pre-training phase, we freeze the visual modality encoder and fine-tune only the textual and signal modality encoders. This strategy aims to maximize the feature similarity between different modality encoders. Formally, given two modalities $m_A$ and $m_B$ corresponding to the same action, with their encoded feature vectors denoted as $f_A$ and $f_B$, respectively, we define the loss function as 
\begin{equation}
    \mathcal{L}_{AB} = 1 - \text{CosSim}(f_A, f_B) = 1 - \frac{f_A \cdot f_B}{\|f_A\| \cdot \|f_B\|}
\end{equation}

The overall loss function is defined as 
\begin{equation}
    \mathcal{L} \triangleq \alpha_{vi}\mathcal{L}_{vi} + \alpha_{vt}\mathcal{L}_{vt} + \alpha_{ti}\mathcal{L}_{ti}\
    \label{eq:loss}
\end{equation}

During inference, we extract features from the input signal and compute their similarity with those of the textual labels. The label with the highest similarity score is assigned as the predicted class of the signal.

\begin{table*}[!t]
\centering
\caption{Human Activity Recognition Results}
\small
\begin{tabular}{c|ccc|ccc|ccc|ccc|ccc|ccc}
\hline
       Label                         & \multicolumn{3}{c|}{SVM}  & \multicolumn{3}{c|}{Random Forest} & \multicolumn{3}{c|}{MLP}  & \multicolumn{3}{c|}{CNN}  & \multicolumn{3}{c|}{LSTM} & \multicolumn{3}{c}{Ours} \\ \cline{2-19} 
                                & P       & R      & F1     & P          & R         & F1        & P       & R      & F1     & P       & R      & F1     & P      & R      & F1 & P          & R         & F1    \\ \hline
0                             & 0.97    & 0.89   & 0.93   & 0.97       & 0.89      & 0.93      & 0.97    & 0.90   & 0.93   & 0.98   &   0.86    &  0.92   & 0.97   & 0.91   & 0.94 &  0.99   &   0.94   &   0.96\\
1                           & 0.79    & 0.87   & 0.83   & 0.85       & 0.87      & 0.86      & 0.80    & 0.85   & 0.83   & 0.85   &   0.87   &   0.86   & 0.85   & 0.85   & 0.85  & 0.87  &    0.89   &   0.88\\
2              & 0.68    & 0.33   & 0.44   & 0.72       & 0.60      & 0.66      & 0.53    & 0.53   & 0.53   & 0.67  &    0.49   &   0.56   & 0.74   & 0.58   & 0.65   & 0.68  &    0.63    &  0.65  \\
3                            & 0.45    & 0.45   & 0.45   & 0.53       & 0.35      & 0.42      & 0.46    & 0.40   & 0.43   & 0.50  &    0.43   &   0.46   & 0.47   & 0.41   & 0.44 &  0.59  &    0.47 &    0.52 \\
4                             & 0.00    & 0.00   & 0.00   & 0.00       & 0.00      & 0.00      & 0.23    & 0.19   & 0.21   & 1.00   &   0.06     & 0.12   & 0.00   & 0.00   & 0.00 &  0.00   &   0.00   &   0.00\\
5                    & 0.69    & 0.19   & 0.30   & 0.38       & 0.28      & 0.32      & 0.38    & 0.26   & 0.31   & 0.38   &   0.16  &    0.22   & 0.26   & 0.47   & 0.34  & 0.66    &  0.81  &    0.73  \\
6             & 0.81    & 0.73   & 0.77   & 0.84       & 0.77      & 0.80      & 0.76    & 0.79   & 0.78   & 0.86   &   0.77   &   0.82   & 0.84   & 0.76   & 0.80  & 0.84    &  0.85  &    0.85\\
7                 & 0.63    & 0.86   & 0.73   & 0.63       & 0.84      & 0.72      & 0.70    & 0.75   & 0.72   & 0.68   &   0.87   &   0.77   & 0.72   & 0.79   & 0.75 &  0.78    &  0.80   &   0.79 \\
8              & 0.72    & 0.44   & 0.55   & 0.67       & 0.52      & 0.59      & 0.59    & 0.49   & 0.54   & 0.57  &    0.55   &   0.56   & 0.71   & 0.52   & 0.60   & 0.68  &    0.59   &   0.63\\
9                 & 0.62    & 0.41   & 0.49   & 0.60       & 0.44      & 0.51      & 0.50    & 0.46   & 0.48   & 0.57   &   0.51   &   0.54   & 0.44   & 0.63   & 0.52 &  0.63   &   0.83    &  0.72\\
10                 & 0.00    & 0.00   & 0.00   & 0.67       & 0.17      & 0.27      & 0.15    & 0.12   & 0.14   & 0.75   &   0.12  &    0.21   & 0.29   & 0.21   & 0.24 &  0.67   &   0.58    &  0.62\\
11 & 0.80    & 0.60   & 0.69   & 0.74       & 0.70      & 0.72      & 0.75    & 0.75   & 0.75   & 0.62   &   0.25   &   0.36  & 0.00   & 0.00   & 0.00 &  0.71    &  0.60  &    0.65 \\
12       & 0.85    & 0.90   & 0.88   & 0.84       & 0.83      & 0.84      & 0.83    & 0.90   & 0.86   & 0.75    &  0.81    &  0.78   & 0.80   & 0.81   & 0.81 & 0.83   &   0.85   &   0.84 \\
13         & 0.74    & 0.84   & 0.79   & 0.71       & 0.77      & 0.74      & 0.80    & 0.65   & 0.71   & 0.81  &    0.84    &  0.83   & 0.49   & 0.77   & 0.60 & 0.87   & 0.87   &   0.87  \\ \hline
macro avg                       & 0.63    & 0.54   & 0.56   & 0.65       & 0.57      & 0.60      & 0.60    & 0.57   & 0.59   & 0.71   &   0.54    &  0.57   & 0.54   & 0.55   & 0.54  & 0.70    &  0.69   &   0.69 \\
weighted avg                    & 0.70    & 0.70   & 0.69   & 0.72       & 0.72      & 0.71      & 0.69    & 0.69   & 0.69   & 0.73   &   0.73    
& 0.72   & 0.72   & 0.71   & 0.71  &  0.78  &    0.78    &  0.78 \\ 
accuracy                        & \multicolumn{3}{c|}{70.27\%} & \multicolumn{3}{c|}{71.91\%}          & \multicolumn{3}{c|}{69.46\%} & \multicolumn{3}{c|}{72.91\%} & \multicolumn{3}{c|}{71.41\%} & \multicolumn{3}{c}{78.11\%} \\ \hline
\end{tabular}
\label{classification}
\end{table*}

\section{Experiments}
\label{Experiments}

\subsection{Experiment Setup}

We conduct experiments using a variety of approaches, including traditional machine learning methods (Support Vector Machine, Random Forest, and Multi-Layer Perceptron), deep learning-based techniques, and the proposed multimodal alignment-based model. The experiments are performed on our proposed dataset, which is partitioned into training and testing sets with an 8:2 split. All experiments are carried out on NVIDIA A100 GPUs.

The quantitative evaluation metrics for the experiments included precision (P), recall (R), and F1-score for activity recognition. For each method, we calculated the precision, recall, and F1-score for each class, as well as the macro-average, weighted average, and overall accuracy.

\subsection{Implementation Details}

The SVM model utilized a radial basis function (RBF) kernel with a regularization parameter of $C = 1$. For the Random Forest method, 100 trees were employed. The MLP was a two-layer fully connected network with 128 and 64 nodes, respectively. The CNN architecture included three convolutional layers followed by two fully connected layers. The LSTM network consisted of three layers, with 12 input features and a hidden state dimension of 128. All models were trained using the Adam \cite{kingmaAdamMethodStochastic2017} optimizer, with a learning rate decaying from $1 \times 10^{-3}$ to $1 \times 10^{-4}$. The hyperparameters in equation \ref{eq:loss} were set to 1.

\begin{figure}[!t]
\centerline{\includegraphics[width=0.9\columnwidth]{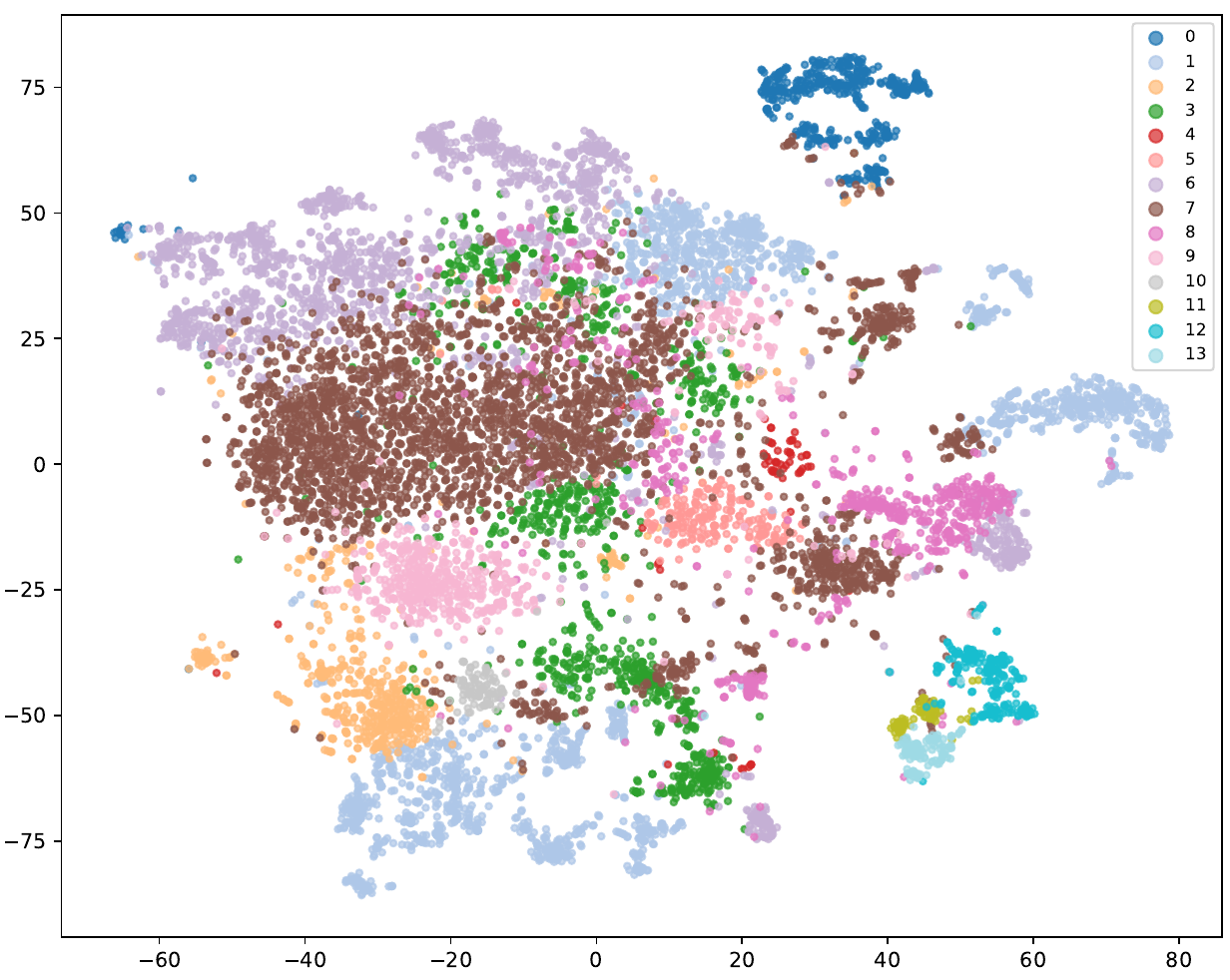}}
\caption{The t-SNE visualization of features extracted by our proposed method. }
\label{tsne}
\Description{The t-SNE visualization of features extracted by our proposed method.}
\end{figure}

\begin{figure}[!t]
\centerline{\includegraphics[width=0.96\columnwidth]{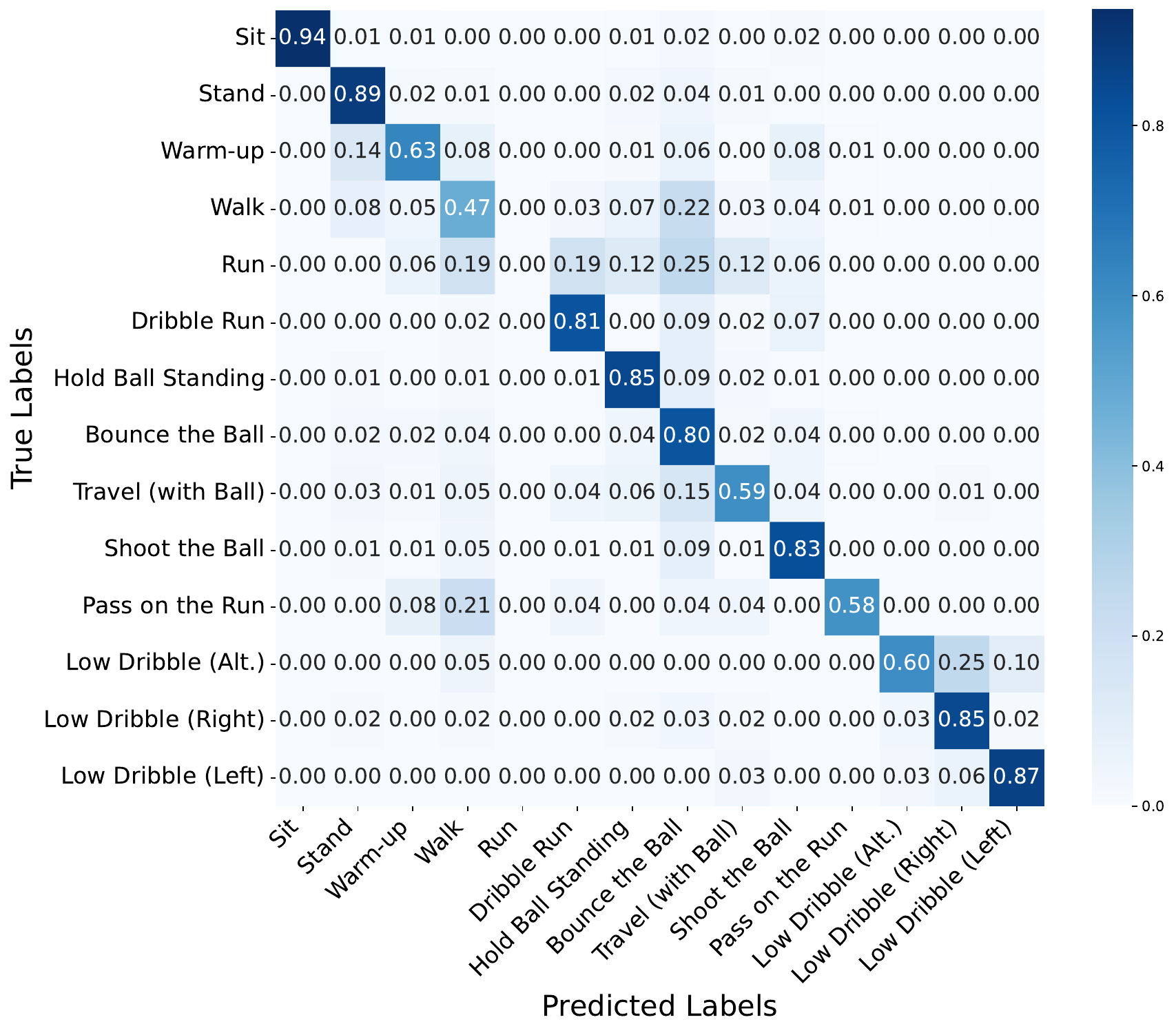}}
\caption{The confusion matrix of classification using our proposed method. }
\label{heatmap}
\Description{The confusion matrix of classification using our proposed method. }
\end{figure}

\subsection{Results and Analysis}
\label{Results}

Table~\ref{classification} presents the experimental results of Human Activity Recognition using SVM, Random Forest, MLP, a simple CNN, and LSTM models. The results indicate that a three-layer CNN achieves an accuracy of 73\%. With our proposed multimodal fusion training strategy, the recognition accuracy can be further improved to over 78\%, underscoring the challenging nature of our dataset for HAR tasks. Figure~\ref{tsne} shows the t-SNE visualization of temporal features extracted from the BasketHAR dataset using our method. Although the decision boundaries between certain activity classes are relatively close—making it difficult to separate samples near the boundaries—features of most samples are well-clustered according to their respective categories, thereby enabling effective classification. 

Figure~\ref{heatmap} depicts the confusion matrix obtained using our classification approach, offering a deeper understanding of the classifier’s performance across different activity categories. The matrix reveals that the classifier encounters significant difficulty in accurately recognizing activities such as ``walking,'' ``running,'' and ``dribbling while running.'' In fact, most methods yield less than 50\% accuracy on these categories, suggesting substantial challenges in achieving precise classification. This is largely due to the high similarity of these running-related activities to other behaviors, and their frequent co-occurrence with actions such as dribbling, passing, and bouncing. Consequently, more expressive and discriminative representations of activity features are required. We encourage researchers to conduct further experiments on this dataset, extract diverse and robust features from both samples and sub-samples, and develop more compelling and effective recognition algorithms.

\section{Potential Applications}
\subsection{Human Activity Recognition}

Our dataset can be employed to train HAR models for both everyday activities and domain-specific actions in the context of basketball. As demonstrated in Section~\ref{Experiments}, models trained on our dataset can achieve an overall accuracy of 78\%, underscoring the inherent complexity and challenge posed by the dataset. The use of more advanced models holds the potential to further enhance the accuracy of human activity recognition.

\subsection{Prompt Optimization for Training Analysis Report Generation}

 The advancement of large language models (LLMs) has enabled the generation of domain-specific analytical reports by leveraging the expert knowledge acquired during their pretraining phase \cite{gaoMotionSignalsInsights2025}. Given that prompt design plays a critical role in determining the quality of generated reports, and our dataset provides a prompt optimization pathway closely aligned with real-world application scenarios, users can iteratively refine their prompts using the rich multimodal data provided by our dataset to obtain insightful, expert-level analytical reports that meet their expectations.

\begin{figure}[!t]
\centerline{\includegraphics[width=\columnwidth]{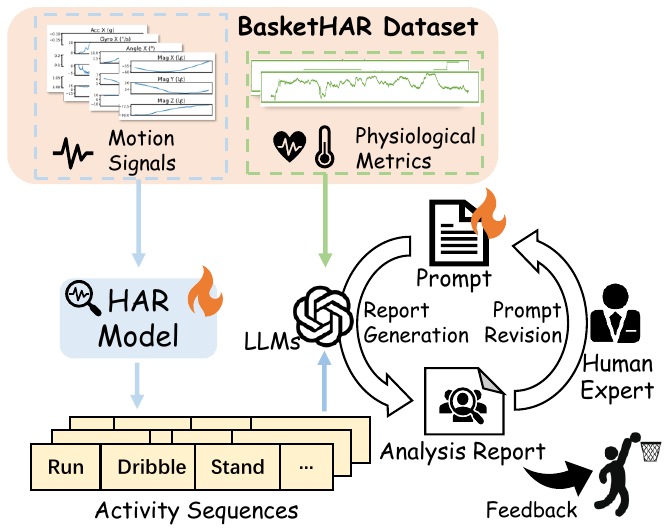}}
\caption{A human-in-the-loop prompt optimization framework based on our dataset.}
\label{use_case}
\Description{A human-in-the-loop prompt optimization framework based on our dataset.}
\end{figure}

\begin{figure}[!t]
\centerline{\includegraphics[width=0.95\columnwidth]{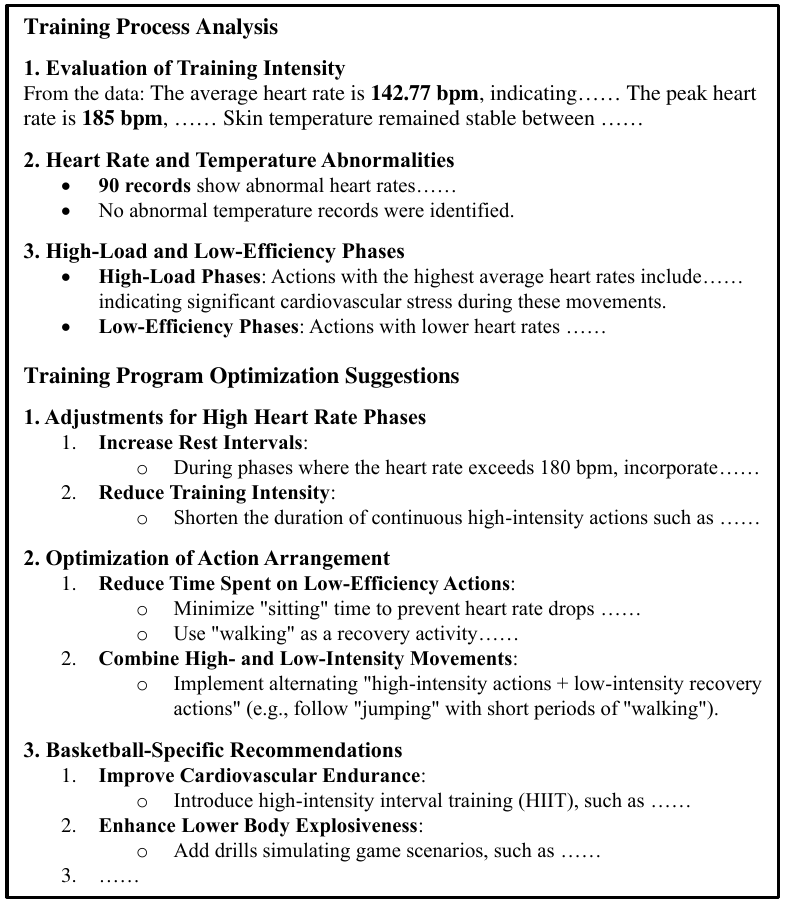}}
\caption{An excerpt of a sports analysis report generated by LLM.}
\label{report}
\Description{}
\end{figure}

As illustrated in Figure~\ref{use_case}, we propose a human-in-the-loop prompt optimization framework based on our dataset. After translating motion signals into behavior sequences interpretable by LLMs via Human Activity Recognition (HAR), the LLM generates specialized analytical reports grounded in both motion content and physiological indicators. Human experts then iteratively refine the prompts fed to the LLM to ensure the quality and relevance of the outputs. Utilizing the state-of-the-art ChatGPT-4o model and a meticulously designed templated prompt structure, we generated a high-quality analysis report and feedback recommendations based on motion data and physiological metrics. An excerpt of the resulting sports analysis report is shown in Figure~\ref{report}.

\section{Conclusion and Future Work}

This paper presents and details the BasketHAR dataset, which comprises information on 14 distinct activities, encompassing both everyday actions and specialized basketball training movements, recorded using a variety of sensors. We provide essential insights into the dataset’s construction and annotation procedures, along with a baseline method based on multimodal alignment. Experimental results demonstrate that the dataset poses a challenging recognition task, validating its utility for advancing HAR research. Finally, we illustrate the dataset’s potential for more in-depth applications, such as the generation of personalized sports performance reports. 

We hope that researchers will incorporate this dataset into their own work and explore it from novel and diverse perspectives. Future plans include continued data collection with sensors placed at multiple body locations to further increase the dataset’s scale and diversity.

\section*{Ethical Statement}
We have obtained authorization to collect and use signals and videos for research purposes. All identifiable information, such as names, faces or markers, has been anonymized. The signals are strictly used for the objectives outlined in this study, with no alternative purposes. Extensive security measures have been implemented to safeguard the database from unauthorized access, ensuring responsible data use. Additionally, we support ongoing ethical reviews and community consultations to continually refine the ethical and implementation guidelines of the framework.

\bibliographystyle{ACM-Reference-Format}
\bibliography{baskethar}

\end{document}